\documentclass[sigconf]{acmart}
\AtBeginDocument{%
  }

\setcopyright{acmlicensed}

\copyrightyear{2026}
\acmYear{2026}
\setcopyright{cc}
\setcctype{by}
\acmConference[MM '26]{Proceedings of the 34th ACM International Conference on Multimedia}{November 10--14, 2026}{Rio de Janeiro, Brazil}
\acmBooktitle{Proceedings of the 34th ACM International Conference on Multimedia (MM '26), November 10--14, 2026, Rio de Janeiro, Brazil}
\acmDOI{10.1145/3767308.3835680}
\acmISBN{979-8-4007-2213-4/2026/11}

\usepackage{subcaption}
\usepackage{multirow}
\usepackage{pifont}
\usepackage{booktabs, colortbl, xcolor}
\usepackage{balance}

\begin{document}

\title{LLaVAFlow: Preserving Latent Alignment Flow for Parameter-Efficient Multimodal Fine-Tuning}

\author{Muyao Yuan}
\authornote{Both authors contributed equally to this research.}
\orcid{0009-0003-2721-6266}
\affiliation{
  \department{MOEKLINNS, and School of Computer Science and Technology}
  \institution{Xi'an Jiaotong University}
  \city{Xi'an}
  \country{China}
}
\email{yuanmuyao@stu.xjtu.edu.cn}

\author{Muyan Jiao}
\authornotemark[1]
\orcid{0000-0001-6614-1664}
\affiliation{
  \department{MOEKLINNS, and School of Computer Science and Technology}
  \institution{Xi'an Jiaotong University}
  \city{Xi'an}
  \country{China}
}
\email{jmy0406@stu.xjtu.edu.cn}

\author{Jiangyong Ying}
\orcid{0000-0002-8819-8538}
\affiliation{
  \department{E-surfing Vision Technology Co., Ltd}
  \institution{China Telecom}
  \city{Hangzhou}
  \country{China}
}
\email{yingjiangyong@chinatelecom.cn}

\author{Weizhan Zhang}
\authornote{Corresponding author.}
\orcid{0000-0003-0330-5435}
\affiliation{
  \department{MOEKLINNS, and School of Computer Science and Technology}
  \institution{Xi'an Jiaotong University}
  \city{Xi'an}
  \country{China}
}
\email{zhangwzh@xjtu.edu.cn}

\author{Yuanhong Zhang}
\orcid{0009-0006-0281-2987}
\affiliation{
  \department{MOEKLINNS, and School of Computer Science and Technology}
  \institution{Xi'an Jiaotong University}
  \city{Xi'an}
  \country{China}
}
\email{yuanhongzhang@stu.xjtu.edu.cn}

\author{Lan Ma}
\orcid{0009-0007-9842-685X}
\affiliation{
  \institution{China Telecom}
  \city{Xi'an}
  \country{China}
}
\email{malan@chinatelecom.cn}

\author{Yuan Gao}
\orcid{0009-0004-7570-1677}
\affiliation{
  \institution{China Telecom}
  \city{Xi'an}
  \country{China}
}
\email{gaoy97@chinatelecom.cn}

\author{Haipeng Du}
\orcid{0000-0002-1120-7096}
\affiliation{
  \department{MOEKLINNS, and School of Computer Science and Technology}
  \institution{Xi'an Jiaotong University}
  \city{Xi'an}
  \country{China}
}
\email{duhaipeng@xjtu.edu.cn}

\renewcommand{\shortauthors}{Muyao Yuan et al.}

\begin{abstract}
While Multimodal Large Language Models (MLLMs) exhibit strong generalization, visual instruction tuning for downstream tasks inevitably causes catastrophic forgetting, impairing overall generalization. 
While existing methods regulate weight updates to reduce forgetting, they overlook the fundamental cross-modal alignment in MLLMs. 
Based on prior work and our observations, we argue that cross-modal alignment is implicitly captured in the information-compression trajectory.
To preserve the alignment flow embedded in the trajectory, we propose LLaVAFlow, an information-theoretic distillation framework.
First, we compress the mutual information between the extracted relations and MLLM embeddings, encouraging a learnable module to produce a refined alignment flow that benefits downstream tasks.
Second, we maximize the mutual information between the extracted alignment flows of the pretrained and fine-tuned MLLMs, enabling the transfer of compact alignment information.
Extensive experiments show that LLaVAFlow is an effective plug-and-play framework that preserves alignment flow and enhances both downstream performance and generalization.
\end{abstract}




\maketitle

\section{Introduction}
Multimodal large language models (MLLMs) exhibit strong generalization, enabling effective transfer across modalities and tasks~\cite{liu2024improved,chen2024internvl,li2024llava}. 
To enhance MLLM performance on specific downstream tasks, visual instruction tuning~\cite{liu2023visual} is widely adopted.
However, visual instruction tuning introduces significant generalization risks, as it disrupts the original feature space of the pretrained MLLM.
Extensive studies~\cite{li2024revisiting,dou2024loramoe, han2025slim} have shown that fine-tuning can lead to catastrophic forgetting, and this problem becomes even more severe in MLLMs because of their complex architectures, larger modality gaps, and stronger task interference, making it particularly challenging to balance general knowledge retention with improvements on downstream tasks~\cite{shen2024multimodal,sung2023ecoflap,zhai2024investigating}.

\begin{figure}
    \centering
    \includegraphics[width=\linewidth]{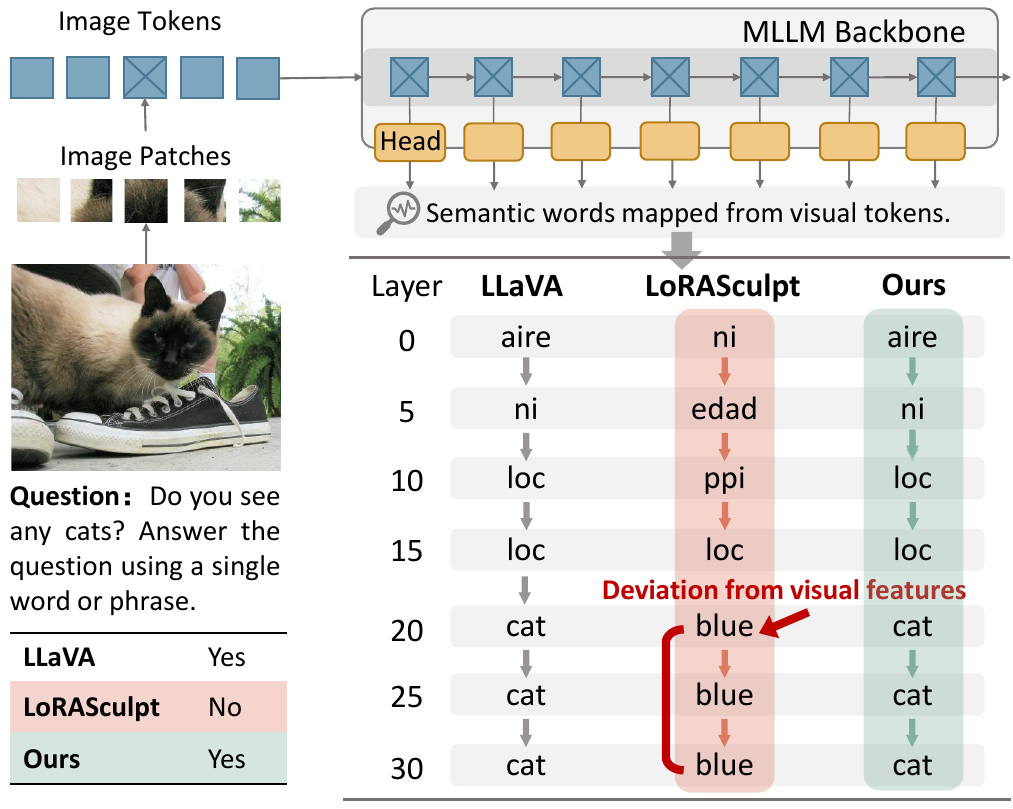}
    \caption{\textbf{Motivation.} We map visual tokens to semantic words along the forward pass using Logit Lens~\cite{nostalgebraist2020logitlens}. Existing methods~\cite{liang2025lorasculpt} disrupt the alignment within the feature flow, leading to incorrect predictions. By preserving the alignment flow, we effectively mitigate forgetting.}
    \label{fig:motivation}
\end{figure}

To mitigate generalization issues, existing approaches~\cite{yu2024language,zhu2024model,liang2025lorasculpt} constrain weight updates through regularization and pruning to limit harmful information and reduce conflicts with the model’s prior general knowledge. However, these methods operate solely on weight-level analysis and largely ignore multimodal information. A central characteristic of MLLMs is their cross-modal alignment, which has often been overlooked during downstream transfer~\cite{zhao2024aligngpt,huang2025deciphering}. 
As a result, existing fine-tuning methods inadvertently disrupt the alignment between visual and textual features, a key cause of catastrophic forgetting, in which the model fails to ground images to their correct semantics, degrading its comprehension and reasoning capabilities.

Since cross-modal alignment directly affects downstream performance, understanding how it forms is essential for preserving it. Existing studies~\cite{neo2024towards,zhang2025cross} show that in MLLMs, visual and textual embeddings align and fuse 
mainly within the LLM backbone.
Specifically, as illustrated in Figure~\ref{fig:motivation}, when the image embedding of a \textit{cat} patch is processed, we capture its semantic evolution by applying the LM head to the hidden states (known as Logit Lens~\cite{nostalgebraist2020logitlens}). The representation transitions from an initially non-semantic state into the explicit text token: \textit{cat}. However, after fine-tuning, this trajectory is disrupted and the model can no longer arrive at the correct semantic interpretation. This observation suggests that mitigating catastrophic forgetting of cross-modal alignment amounts to preserving the underlying semantic trajectory. We term this trajectory the \textbf{alignment flow}, through which visual features are progressively grounded into textual semantics.

To preserve the alignment flow during downstream fine-tuning, a natural approach is to distill it from the pretrained model into the fine-tuned one.
However, this poses two key challenges. 
First, the alignment flow is inherently latent~\cite{neo2024towards}, and the internal representation of the LLM backbone contains significant noise irrelevant to both modality alignment and downstream tasks. 
Second, existing distillation methods~\cite{shu2025llava,cai2025llava,feng2025align} typically rely on rigid mimicry of final output distributions or individual features, which is inadequate for preserving modality-dependent feature flows and impedes the model's ability to acquire new task-specific knowledge.

This motivates the need for a method that \textbf{\textit{1) extracts alignment information from the feature trajectory}} and \textbf{\textit{2) enables its effective transfer between models}}, mitigating catastrophic forgetting while facilitating the acquisition of new knowledge.

Building on this insight, we propose LLaVAFlow, an information-theoretic distillation framework. To tackle the first challenge, we introduce a novel Alignment Flow Module (AFM) that models the alignment flow via inter-embedding relations, enabling efficient extraction of modality alignment information. By compressing the mutual information between the extracted relations and the original MLLM embeddings, the AFM produces a refined alignment flow beneficial for downstream tasks. To address the second challenge, we maximize the mutual information between the AFM-extracted alignment flows of the pretrained and fine-tuned MLLMs. Unlike rigid mimicry, this objective leverage mutual information to preserve structural knowledge and transfer compact alignment information.
Notably, these two objectives jointly realize the information bottleneck principle, preserving alignment-relevant information while discarding irrelevant input.

Extensive experiments on VQA and Captioning tasks demonstrate that LLaVAFlow is an effective plug-and-play framework, achieving high downstream performance while mitigating generalization loss in MLLMs through alignment flow preservation.
Our contributions are threefold:




\ding{182} \textbf{A Novel Perspective on Alignment Flow.} We introduce alignment flow as a new lens for preserving pretrained knowledge, encoding key cross-modal information within feature propagation.

\ding{183} \textbf{Information-Theoretic Distillation Framework.} We propose LLaVAFlow, which extracts and transfers alignment information between pretrained and fine-tuned MLLMs, mitigating catastrophic forgetting while effectively acquiring new knowledge.

\ding{184} \textbf{Comprehensive Experimental Validation.} Extensive experiments on VQA and captioning demonstrate that LLaVAFlow effectively preserves alignment flow, improves downstream performance, and mitigates generalization loss.

\section{Related Works}

\subsection{Catastrophic Forgetting}
Deep learning models often suffer from catastrophic forgetting, where acquiring new knowledge causes the loss of previously learned information~\cite{kirkpatrick2017overcoming}. Various continual learning strategies have been proposed to tackle this problem, usually classified as rehearsal-based~\cite{chaudhry2019tiny,lopez2017gradient}, regularization-based~\cite{kirkpatrick2017overcoming,liu2018rotate}, or architecture-based methods~\cite{razdaibiedina2023progressive,wang2023rehearsal}.
For large foundation models, the challenge has shifted to preserving the general knowledge encoded in the pre-trained model after task-specific fine-tuning. 
Many methods have been proposed to mitigate catastrophic forgetting in LLM~\cite{yu2024language,panigrahi2023task,yang2024corda}. 
However, they show limited effectiveness on MLLMs due to the complex architecture and non-uniform weight distributions resulting from their cross-modal nature~\cite{zhu2024model}.
For catastrophic forgetting in MLLMs, the EMT~\cite{zhai2024investigating} evaluates LLaVA and InstructBLIP using classification task performance but does not examine broader multimodal tasks. 
Model Tailor~\cite{zhu2024model} offers a post-training adjustment method that preserves most pre-trained parameters while selectively updating a small fraction of weights.
LoRASculpt~\cite{liang2025lorasculpt} mitigates conflicts between incremental and original weights and prunes harmful updates during training, effectively maintaining performance on previous tasks while accommodating new ones.
However, these methods primarily focus on weight updates and overlook the loss of modality alignment during fine-tuning. To address this, we propose LLaVAFlow, an orthogonal approach that preserves essential alignment information in feature propagation, effectively mitigating catastrophic forgetting in MLLMs.

\subsection{Knowledge Distillation}
Knowledge distillation (KD) aims to transfer knowledge from a large, well-trained teacher model to a smaller, lightweight student model~\cite{hinton2015distilling,yuan2024adaptive}. Early distillation methods for VLMs are primarily designed prior to the foundation model era~\cite{fang2021compressing,wang2023efficientvlm,yan2026vldus}. Although these works provide valuable insights into multimodal knowledge transfer, they are inadequate for capturing the complex cross-modal alignment inherent in modern MLLMs~\cite{feng2025align}.
Recent efforts have begun to explore distillation tailored specifically for MLLMs. 
VLM-KD~\cite{zhang2024vlm} leverages LLaVA-Next to generate textual prompts and employs contrastive learning to enhance long-tail recognition in vision models. 
LLaVA-MoD~\cite{shu2025llava} introduces a progressive distillation framework that equips the small-scale MLLM (s-MLLM) with a sparse Mixture-of-Experts (MoE) architecture and transfers knowledge from the large-scale MLLM (l-MLLM) through KL-based mimic learning followed by preference optimization. 
LLaVA-KD~\cite{cai2025llava} proposes a multimodal distillation framework that transfers cross-modal representations and visual token relations via MDist and RDist losses, improving both alignment quality and multimodal understanding in s-MLLMs through a three-stage training scheme. 
Align-KD~\cite{feng2025align} introduces a cross-modal distillation method that explicitly transfers shallow-layer alignment knowledge and vision-to-text projection capabilities from teacher VLMs, enhancing student performance without increasing model size.
Despite these advances, existing methods predominantly focus on transferring knowledge from l-MLLMs to s-MLLMs while overlooking the alignment flow information, which is crucial for preserving general knowledge during task-specific fine-tuning and mitigating catastrophic forgetting. In contrast, our method addresses this gap by explicitly modeling and preserving alignment flow throughout the distillation process.

\begin{figure*}
    \centering
    \includegraphics[width=0.9\linewidth]{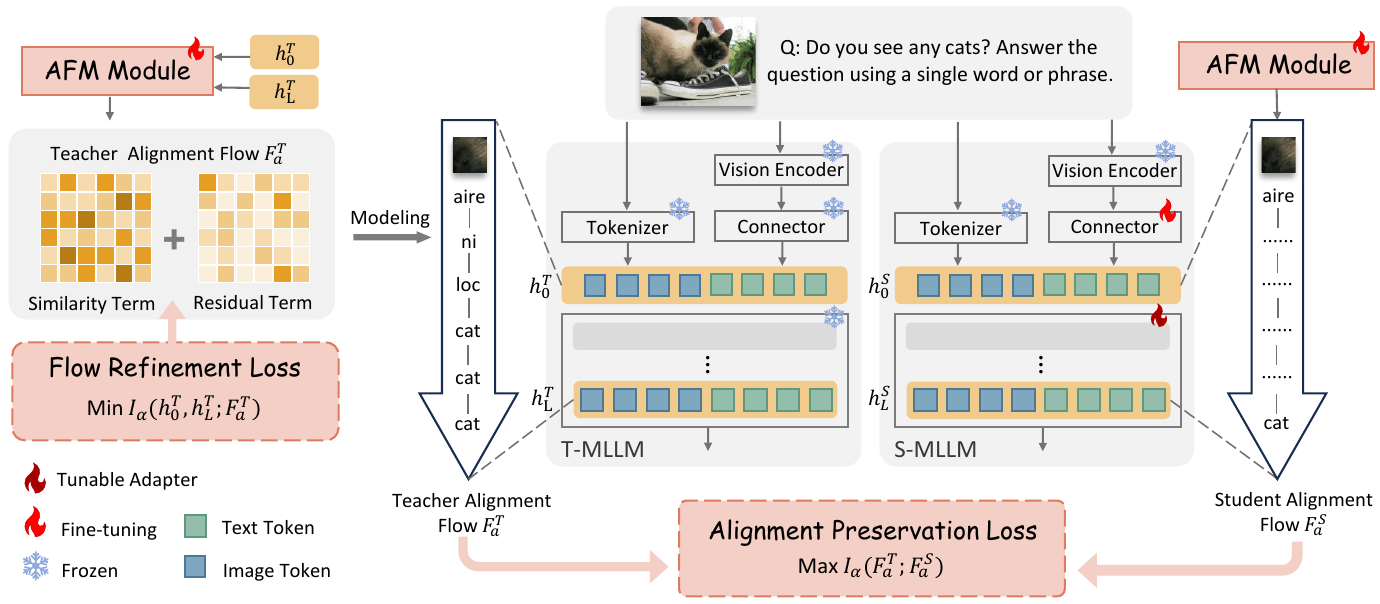}
    \caption{\textbf{Overview of LLaVAFlow.} LLaVAFlow is a distillation framework that effectively transfers alignment flow from a pretrained MLLM to its fine-tuned counterpart from an information-theoretic perspective. The framework comprises an Alignment Flow Module (AFM) and an Information Bottleneck Distillation (IBD) objective that jointly regulate MLLM fine-tuning: the Flow Refinement Loss drives the AFM to capture refined alignment flow, while the Alignment Preservation Loss effectively transfers it to the student. The T-MLLM (teacher) and S-MLLM (student) share the same AFM parameters.}
    \label{fig:overview}
\end{figure*}

\section{Preliminaries}
\subsection{Matrix-Based Rényi’s $\alpha$-entropy}
Rooted in information theory, Rényi’s $\alpha$-entropy generalizes Shannon entropy and provides a flexible measure for quantifying both single-variable information and interactions among variables. It is defined for a continuous random variable $X$ with probability density function $p(x)$ over a finite set $\mathcal{X}$ as:
\begin{equation}
H_{\alpha}(X) = \frac{1}{1 - \alpha} \log \int_{\mathcal{X}} p^{\alpha}(x)\, dx,
\end{equation}
where $H_{\alpha}(X)$ recovers to Shannon entropy as $\alpha \to 1$. However, its computation depends on the probability density function $p(x)$, which is generally inaccessible in MLLM learning. Matrix-Based Rényi’s $\alpha$-entropy overcomes this limitation by directly estimating entropy from data through the eigenspectrum of a Gram matrix in a reproducing kernel Hilbert space (RKHS).

For a single variable, let $\kappa: \mathcal{X} \times \mathcal{X} \to \mathbb{R}$ be a positive, infinitely divisible kernel~\cite{bhatia2006infinitely}. Given sampled data $\{x_i\}_{i=1}^n \subset \mathcal{X}$, where each $x_i$ is a real-valued scalar or vector, the Gram matrix $K$ is constructed with entries $K_{ij} = \kappa(x_i, x_j)$. 
The matrix-based Rényi's $\alpha$-entropy is defined on a positive semi-definite matrix $\mathbf{A}$ with unit trace as:
\begin{equation}
S_{\alpha}(\mathbf{A}) = \frac{1}{1-\alpha} \log \big( \mathrm{tr}(\mathbf{A}^\alpha) \big) = \frac{1}{1-\alpha} \log \left( \sum_{i=1}^{n} \lambda_i^\alpha(\mathbf{A}) \right),
\label{eq:matrix-based-entropy}
\end{equation}
where $\lambda_i(\mathbf{A})$ denotes the $i$-th eigenvalue of $\mathbf{A}$. 
In practice, $\mathbf{A}$ is constructed via a normalized kernel matrix:
\begin{equation}
\mathbf{A}_{ij} = \frac{1}{n} \frac{K_{ij}}{\sqrt{K_{ii} K_{jj}}},
\label{eq:normalized_kernel_matrices}
\end{equation}
which preserves positive semi-definiteness and satisfies $\mathrm{tr}(\mathbf{A}) = 1$.

For multiple variables, let $\kappa_1: \mathcal{X}_1 \times \mathcal{X}_1 \to \mathbb{R}, \dots, \kappa_L: \mathcal{X}_L \times \mathcal{X}_L \to \mathbb{R}$ be positive, infinitely divisible kernels, and let $\{x_i^1, \dots, x_i^L\}_{i=1}^n \subset \mathcal{X}_1 \times \cdots \times \mathcal{X}_L$ denote a set of $n$ samples. The matrix-based Rényi $\alpha$-order joint entropy of these $L$ variables is defined as
\begin{equation}
S_{\alpha}(\mathbf{A}_1, \dots, \mathbf{A}_L) = S_{\alpha} \Bigg( \frac{\mathbf{A}_1 \circ \cdots \circ \mathbf{A}_L}{\mathrm{tr}(\mathbf{A}_1 \circ \cdots \circ \mathbf{A}_L)} \Bigg),
\label{eq:matrix-based-joint-entropy}
\end{equation}
where $\mathbf{A}_1, \dots, \mathbf{A}_L$ are the normalized kernel matrices for each variable, and $\circ$ denotes the Hadamard product.

\subsection{Matrix-Based Mutual Information}
Since mutual information can be expressed in terms of entropy, the mutual information between $\mathbf{A}_1$ and $\mathbf{A}_2$, denoted as $I_\alpha(\mathbf{A}_1; \mathbf{A}_2)$, is given by
\begin{equation}
I_\alpha(\mathbf{A}_1; \mathbf{A}_2) = S_\alpha(\mathbf{A}_1) + S_\alpha(\mathbf{A}_2) - S_\alpha(\mathbf{A}_1, \mathbf{A}_2).
\label{eq:matrix-based-MI}
\end{equation}
This matrix-based approach avoids high-dimensional probability density estimation, providing a more accurate and computationally efficient formulation for mutual information~\cite{yu2019multivariate}.

\section{Method}
\subsection{Overview of LLaVAFlow}
The overview of LLaVAFlow is presented in Figure~\ref{fig:overview}. LLaVAFlow adopts a distillation paradigm, where a frozen pretrained MLLM serves as the teacher, and a parameter-efficiently fine-tuned model trained from it acts as the student. The framework comprises an \textbf{Alignment Flow Module (AFM)} and an \textbf{Information Bottleneck Distillation (IBD)} that jointly regulate the fine-tuning of the MLLM.

The AFM captures multimodal alignment during feature propagation. Specifically, based on data processing inequality (DPI), it takes the embeddings before the first layer and after the final layer of the LLM backbone as inputs, and extracts an alignment flow that characterizes the evolution of multimodal representations across the backbone. The AFM is attached to both the teacher and student models with shared parameters.

Based on this alignment flow, we formulate distillation as an information bottleneck objective composed of two coupled terms. The \textbf{Flow Refinement Loss} serves as the compression term on the AFM of the teacher model by minimizing the mutual information between the alignment flow and the LLM backbone embeddings, thereby suppressing noise in the extracted flow. The \textbf{Alignment Preservation Loss} acts as the distillation term by maximizing the mutual information between the alignment representations of the pretrained and fine tuned MLLMs, which transfers alignment knowledge from the pretrained model and prevents the degradation of flow alignment during adaptation.

\subsection{Alignment Flow Module (AFM)}
\label{sec:AFM}
The Alignment Flow Module (AFM) is designed to capture the evolution of multimodal alignment within a model. 
Intuitively, valuable alignment information can be extracted from the trajectory of multimodal embeddings.
Let $\{h_l\}_{l=0}^L$ represent the multimodal embeddings at each layer of an $L$-layer LLM backbone ($h_0$ denotes the embeddings before the first layer).
In principle, the AFM can be formulated as a function of all layer embeddings to obtain the alignment flow $F_a$:
\begin{equation}
F_a = f_{AFM}(h_0, h_1, \dots, h_L).
\end{equation}
However, performing computations across all hidden embeddings is impractical. Since each layer in the LLM backbone processes inputs exclusively from its immediate predecessor, the model effectively forms a Markov chain across its layers~\cite{tishby2015deep}.
Let $Y$ represent the output of the multimodal task. According to the data processing inequality (DPI)~\cite{cover1999elements,tishby2015deep,kawaguchi2023does}, the mutual information between the representation at each layer and the final output must satisfy:
\begin{equation}
I(h_0; Y) \ge I(h_1; Y) \ge \dots \ge I(h_L; Y).
\end{equation}
This implies that as representations propagate through the layers, the quantity of task-relevant information is non-increasing.
Empirically, after training, the final-layer embedding $h_L$ retains nearly all task-relevant information from the input $h_0$. Consequently, the AFM can effectively capture the core multimodal alignment using only the input and output embeddings:
\begin{equation}
F_a = f_{AFM}(h_0, h_L),
\end{equation}
which significantly reduces computational cost while preserving essential alignment information.

Specifically, the AFM learns a relation between $h_0$ and $h_L$ to characterize the alignment flow. We model this relation using a similarity term and further introduce an adaptive attention mechanism as a residual compensation to refine it. Given $h_0, h_L \in \mathbb{R}^{B\times N \times D}$, where $B$, $N$, and $D$ denote the batch size, the number of multimodal tokens, and the embedding dimension, respectively, the AFM computes a relation map $F_a \in \mathbb{R}^{B\times N \times N}$, where each entry reflects the relevance between an input token and an output token. Let $\tilde{h}_L = \mathrm{LayerNorm}(h_L)$ and $\tilde{h}_0 = \mathrm{LayerNorm}(h_0)$. The base similarity is computed by the dot product of the normalized embeddings $\tilde{h}_L \tilde{h}_0^\top$. To refine this relation, we introduce a residual component implemented with a scaled dot-product attention mechanism: $Q = W_Q \tilde{h}_L, \; K = W_K \tilde{h}_0$, where $W_Q, W_K \in \mathbb{R}^{D\times D}$ are learnable projection matrices. The final alignment flow is computed as
\begin{equation}
F_a = f_{AFM}(h_0, h_L) = \tilde{h}_L \tilde{h}_0^\top + QK^\top / \sqrt{D}.
\end{equation}
In this way, the AFM can adaptively adjust how the relation between $h_0$ and $h_L$ is modeled, providing a robust and effective mechanism to capture alignment information in the flow.

\subsection{Information Bottleneck Distillation (IBD)}

The goal of Information Bottleneck Distillation (IBD) is to transfer the refined alignment flow from the teacher to the student. Specifically, IBD addresses two key questions: how to capture critical alignment information from the teacher’s feature propagation and how to effectively preserve it in the student during fine-tuning. Accordingly, we propose the Flow Refinement Loss and the Alignment Preservation Loss under the information bottleneck principle.

\paragraph{Flow Refinement Loss} 
While the AFM described in Sec.~\ref{sec:AFM} extracts alignment flow from the embeddings before the first layer and after the final layer, these embeddings contain substantial redundancy beyond the alignment information useful for the downstream task. To address this issue, we adopt the compression principle of the information bottleneck to refine the alignment flow. 

Given the teacher MLLM’s input embeddings $h_0^T$ and output embeddings $h_L^T$, the AFM produces the corresponding alignment flow $F_a^T$. Our objective is to regulate this flow by minimizing the mutual information $I_\alpha(h_0^T, h_L^T; F_a^T)$. Here, $I_\alpha(\cdot;\cdot)$ denotes the $\alpha$-Rényi mutual information, which quantifies the amount of information shared between the backbone embeddings and the extracted alignment flow. This objective encourages $F_a^T$ to discard redundant information from the embeddings while retaining the essential structure of multimodal alignment. In other words, this can be formulated as a loss to supervise the learning of AFM:
\begin{equation}
\mathcal{L}_{FR} = I_\alpha(h_0^T, h_L^T; F_a^T)
\label{eq:loss_fr_IB}
\end{equation}
To make Eq.~\ref{eq:loss_fr_IB} computationally tractable in MLLM training, we leverage the matrix-based mutual information in Eq.~\ref{eq:matrix-based-MI} to reformulate the objective.
\begin{equation}
\begin{aligned}
\mathcal{L}_{FR} &= I_\alpha(h_0^T, h_L^T; F_a^T) \\
&= S_\alpha(\mathbf{G}_0^T, \mathbf{G}_L^T)
+ S_\alpha(\mathbf{G}_F^T)
- S_\alpha(\mathbf{G}_0^T, \mathbf{G}_L^T, \mathbf{G}_F^T),
\end{aligned}
\end{equation}
where $\mathbf{G}_0^T$, $\mathbf{G}_L^T$, and $\mathbf{G}_F^T \in \mathbb{R}^{B \times B}$ denote the normalized Gram matrices defined in Eq.~\ref{eq:normalized_kernel_matrices}, constructed from the embeddings $h_0^T$, $h_L^T$, and the alignment flow $F_a^T$, respectively. Here, $B$ denotes the batch size. In our implementation, all Gram matrices are computed using a degree-$1$ polynomial kernel~\cite{yuan2025infoclip}.
As the teacher model remains frozen during training, the term $S_\alpha(\mathbf{G}_0^T, \mathbf{G}_L^T)$, which depends only on the teacher embeddings, can be omitted.
So $\mathcal{L}_{FR}$ can be reformulated using Eq.~\ref{eq:matrix-based-entropy} and Eq.~\ref{eq:matrix-based-joint-entropy}:
\begin{equation}
\begin{aligned}
\mathcal{L}_{FR}
&= S_\alpha(\mathbf{G}_F^T)
- S_\alpha(\mathbf{G}_0^T, \mathbf{G}_L^T, \mathbf{G}_F^T) \\
&= \frac{1}{1-\alpha}
\left(
\log \sum_{i=1}^{B}\lambda_i^\alpha(\mathbf{G}_F^T)
-
\log \sum_{i=1}^{B}\lambda_i^\alpha(\mathbf{G}_{0LF}^T)
\right),
\end{aligned}
\end{equation}
where $\mathbf{G}_{0LF}^T = (\mathbf{G}_0^T \circ \mathbf{G}_L^T \circ \mathbf{G}_F^T) / \mathrm{tr}(\mathbf{G}_0^T \circ \mathbf{G}_L^T \circ \mathbf{G}_F^T)$, and $\circ$ denotes the Hadamard product.
To avoid the computational cost of eigenvalue calculation, we follow prior work~\cite{yu2019multivariate} by setting $\alpha = 2$, which allows a Frobenius-norm-based approximation of Rényi’s $\alpha$-entropy~\cite{miles2021information,zhang2025infosam}:
\begin{equation}
||\mathbf{G}||_F^2 = \mathrm{tr}(\mathbf{G}\mathbf{G}^\top) = \sum_{i=1}^{B} \lambda_i^2(\mathbf{G}),
\end{equation}
yielding the final computable loss:
\begin{equation}
\mathcal{L}_{FR} = - \log ||\mathbf{G}_F^T||_F^2 + \log ||\mathbf{G}_{0LF}^T||_F^2.
\label{eq:loss_fr_final}
\end{equation}
In this way, the Flow Refinement Loss enforces a compact representation of the alignment flow, reducing noise while preserving its core multimodal structure.

\paragraph{Alignment Preservation Loss}
Under the supervision of the Flow Refinement Loss, the AFM acts as an effective lens for extracting key alignment information from the flow. To preserve this information during fine-tuning, we incorporate the AFM into both teacher and student models with shared parameters, aiming to align the student’s flow $F_a^S$ with the teacher’s $F_a^T$ and thereby guide the alignment process in the student model.

Specifically, we achieve this transfer by maximizing the mutual information between the alignment flows of the pretrained and fine-tuned models. The Alignment Preservation Loss is defined as:
\begin{equation}
\begin{aligned}
\mathcal{L}_{AP} &= -I_\alpha(F_a^T; F_a^S) \\
&= -S_\alpha(\mathbf{G}_F^T) - S_\alpha(\mathbf{G}_F^S) + S_\alpha(\mathbf{G}_F^T, \mathbf{G}_F^S),
\label{eq:loss_ap_IB}
\end{aligned}
\end{equation}
where $\mathbf{G}_F^T, \mathbf{G}_F^S \in \mathbb{R}^{B \times B}$ denote the normalized Gram matrices of $F_a^T$ and $F_a^S$, respectively.
Analogous to~\ref{eq:loss_fr_final}, the computable form of the loss is:  
\begin{equation}
\mathcal{L}_{AP} = \log \|\mathbf{G}_F^T\|_F^2 + \log \|\mathbf{G}_F^S\|_F^2 - \log \|\mathbf{G}_F^{TS}\|_F^2,
\label{eq:loss_ap_final}
\end{equation}
where $\mathbf{G}_F^{TS} = (\mathbf{G}_F^T \circ \mathbf{G}_F^S) / \mathrm{tr}(\mathbf{G}_F^T \circ \mathbf{G}_F^S)$.
The Alignment Preservation Loss is based on mutual information and differs from traditional distillation. Eq.~\ref{eq:loss_ap_final} provides a more relaxed imitation from the student to the teacher, transferring only the most relevant information and mitigating the interference of the distillation loss on the task loss.

\begin{table*}[t]
\centering
\small
\renewcommand{\arraystretch}{0.9} 
\setlength{\tabcolsep}{3.2pt} 
\caption{Comparison with state-of-the-art methods on the visual question answering task ScienceQA and the image captioning task COCO-Caption. \textit{Source} indicates the average performance across upstream datasets, \textit{Target} reports the performance on the fine-tuned dataset, and \textit{Avg} is the mean of \textit{Source} and \textit{Target}.}
\label{tab:main_results}
\begin{tabular}{l||cccc|cc|c||cccc|cc|c}
\toprule
\multirow{2}{*}{Methods} 
& \multicolumn{7}{c||}{ScienceQA} 
& \multicolumn{7}{c}{COCO-Caption} \\
\cmidrule(lr){2-8} \cmidrule(lr){9-15}
& OKVQA & OCRVQA & GQA & TextVQA & \textit{Source} & \textit{Target} & \textit{Avg}
& OKVQA & OCRVQA & GQA & TextVQA & \textit{Source} & \textit{Target} & \textit{Avg} \\
\midrule

\rowcolor{gray!15}
Zero-shot
& 58.00 & 66.20 & 61.94 & 58.21 & 61.09 & 70.22 & 65.65 
& 58.00 & 66.20 & 61.94 & 58.21 & 61.09 & 40.34 &  50.71 \\
\midrule

\multicolumn{15}{l}{\textit{LoRA Rank = 32}} \\
\midrule

LoRA 
& 56.99 & 65.25 & 61.52 & 56.64 & 60.10 & 87.57 & 73.84
& 48.60 & 66.00 & 56.73 & 43.91 & 53.81 & 128.46 & 91.14  \\

\rowcolor{blue!5}
LoRA+Ours 
& 56.94 & 65.65 & 61.19 & 57.07 & \textbf{60.21} & \textbf{89.46} & \textbf{74.84}  
& 52.18 & 67.25 & 59.33 & 51.20 & \textbf{57.49} & \textbf{130.04} & \textbf{93.76} \\

DoRA 
& 55.10 & 61.05 & 59.68 & 54.18 & 57.50 & 88.94 & 73.22 
& 49.69 & 66.50 & 57.62 & 45.81 & 54.90 & 128.26 & 91.58  \\

\rowcolor{blue!5}
DoRA+Ours 
& 56.86 & 65.85 & 61.05 & 56.72 & \textbf{60.12} & \textbf{89.13} & \textbf{74.62} 
& 51.59 & 67.25 & 59.63 & 51.46 & \textbf{57.48} & \textbf{129.97} & \textbf{93.73}  \\

LLaVAMod 
& 57.51 & 65.10 & 61.71 & 56.96 & 60.32 & 87.55 & 73.94  
& 55.34 & 67.10 & 61.39 & 55.26 & 59.77 & 125.99 & 92.88  \\

\rowcolor{blue!5}
LLaVAMod+Ours 
& 57.35 & 66.10 & 61.46 & 57.36 & \textbf{60.57} & \textbf{88.56} & \textbf{74.57} 
& 56.31 & 67.00 & 61.32 & 57.14 & \textbf{60.44} & \textbf{128.13} & \textbf{94.29}  \\

LLaVAKD
& 57.81 & 66.15 & 62.01 & 57.53 & 60.88 & 88.38 & 74.63  
& 55.49 & 67.05 & 61.40 & 55.58 & 59.88 & 126.87 & 93.38 \\

\rowcolor{blue!5}
LLaVAKD+Ours 
& 57.80 & 66.25 & 61.82 & 57.98 & \textbf{60.96} & \textbf{88.73} & \textbf{74.85}   
& 55.91 & 67.15 & 61.08 & 56.71 & \textbf{60.21} & \textbf{129.02} & \textbf{94.62} \\

DARE 
& 56.32 & 64.50 & 61.46 & 55.75 & 59.51 & 87.46 & 73.48
& 46.26 & 65.85 & 54.61 & 39.75 & 51.62 & 128.32 & 89.97 \\

\rowcolor{blue!5}
DARE+Ours 
& 56.21 & 66.15 & 61.13 & 56.95 & \textbf{60.11} & \textbf{89.37} & \textbf{74.74}  
& 51.49 & 66.70 & 59.06 & 50.50 & \textbf{56.94} & \textbf{129.23} & \textbf{93.08} \\

Model Tailor 
& 57.10 & 65.50 & 61.58 & 57.10 & 60.32 & 87.24 & 73.78  
& 49.37 & 66.05 & 57.77 & 44.85 & 54.51 & 128.58 & 91.54 \\

\rowcolor{blue!5}
Model Tailor+Ours 
& 56.93 & 66.10 & 61.43 & 57.11 & \textbf{60.39} & \textbf{89.20} & \textbf{74.80}  
& 52.41 & 67.20 & 60.17 & 52.08 & \textbf{57.96} & \textbf{129.32} & \textbf{93.64}  \\

LoRASculpt 
& 52.78 & 50.60 & 56.65 & 50.29 & 52.58 & 85.26 & 68.92  
& 56.86 & 66.30 & 61.85 & 57.59 & 60.65 & 128.81 & 94.73 \\

\rowcolor{blue!5}
LoRASculpt+Ours 
& 58.00 & 64.95 & 61.70 & 57.54 & \textbf{60.55} & \textbf{85.69} & \textbf{73.12}  
& 57.05 & 67.10 & 61.68 & 57.77 & \textbf{60.90} & \textbf{129.40} & \textbf{95.15}  \\

\midrule

\multicolumn{15}{l}{\textit{LoRA Rank = 64}} \\
\midrule

LoRA 
& 52.14 & 61.05 & 58.90 & 50.60 & 55.67 & 88.49 & 72.08  
& 44.57 & 62.90 & 51.58 & 38.37 & 49.36 & 128.13 & 88.74 \\

\rowcolor{blue!5}
LoRA+Ours 
& 54.11 & 64.15 & 59.60 & 54.65 & \textbf{58.13} & \textbf{89.46} & \textbf{73.79}
& 44.10 & 66.10 & 53.87 & 42.86 & \textbf{51.73} & \textbf{129.33} & \textbf{90.53}  \\

DoRA 
& 51.65 & 62.25 & 52.19 & 50.20 & 54.07 & 88.40 & 71.24 
& 45.70 & 65.70 & 54.82 & 40.32 & 51.64 & 127.86 & 89.75  \\

\rowcolor{blue!5}
DoRA+Ours 
& 52.11 & 63.85 & 57.85 & 51.67 & \textbf{56.37} & \textbf{88.73} & \textbf{72.55}  
& 44.30 & 66.80 & 54.12 & 42.97 & \textbf{52.05} & \textbf{129.20} & \textbf{90.62} \\

LLaVAMod 
& 55.79 & 64.05 & 61.23 & 54.75 & 58.96 & 88.45 & 73.70
& 53.07 & 66.95 & 60.39 & 52.25 & 58.17 & 126.75 & 92.46  \\

\rowcolor{blue!5}
LLaVAMod+Ours 
& 56.28 & 64.35 & 60.50 & 56.65 & \textbf{59.44} & \textbf{88.75} & \textbf{74.10}
& 52.68 & 67.15 & 60.13 & 53.21 & \textbf{58.29} & \textbf{128.39} & \textbf{93.34}  \\

LLaVAKD
& 56.83 & 66.05 & 61.19 & 56.74 & 60.20 & 88.63 & 74.42  
& 55.03 & 66.60 & 61.54 & 55.28 & 59.61 & 125.97 & 92.79  \\

\rowcolor{blue!5}
LLaVAKD+Ours 
& 57.38 & 66.60 & 61.19 & 56.97 & \textbf{60.54} & \textbf{88.80} & \textbf{74.67}  
& 54.79 & 67.40 & 60.88 & 55.39 & \textbf{59.61} & \textbf{127.98} & \textbf{93.80} \\

DARE 
& 50.75 & 58.80 & 58.16 & 48.87 & 54.14 & 88.35 & 71.25  
& 42.17 & 61.55 & 47.42 & 31.34 & 45.62 & 127.50 & 86.56 \\

\rowcolor{blue!5}
DARE+Ours 
& 52.83 & 63.65 & 59.91 & 54.02 & \textbf{57.60} & \textbf{89.32} & \textbf{73.46}  
& 42.98 & 66.00 & 52.31 & 36.42 & \textbf{49.43} & \textbf{128.76} & \textbf{89.09}  \\

Model Tailor 
& 55.00 & 63.55 & 59.83 & 54.15 & 58.13 & 88.45 & 73.29  
& 45.65 & 63.85 & 53.57 & 39.66 & 50.68 & 128.42 & 89.55 \\

\rowcolor{blue!5}
Model Tailor+Ours 
& 54.86 & 65.85 & 60.23 & 55.80 & \textbf{59.18} & \textbf{89.46} & \textbf{74.32}  
& 45.19 & 66.30 & 55.05 & 44.19 & \textbf{52.68} & \textbf{128.97} & \textbf{90.83}  \\

LoRASculpt 
& 53.83 & 49.95 & 57.54 & 49.68 & 52.75 & 85.38 & 69.07   
& 57.10 & 66.15 & 61.88 & 57.80 & 60.73 & 129.06 & 94.90 \\

\rowcolor{blue!5}
LoRASculpt+Ours 
& 57.92 & 65.45 & 61.73 & 57.86 & \textbf{60.74} & \textbf{86.56} & \textbf{73.65} 
& 57.03 & 66.50 & 61.88 & 57.77 & \textbf{60.80} & \textbf{129.74} & \textbf{95.27}  \\

\bottomrule
\end{tabular}
\end{table*}

\paragraph{Overall Loss and Information-Theoretic View}  
To jointly optimize task performance and alignment preservation, our overall training objective combines the downstream task loss with proposed distillation losses. Specifically, we incorporate the task loss (i.e., SFT loss), denoted as $\mathcal{L}_{task}$, together with the Flow Refinement Loss and the Alignment Preservation Loss:
\begin{equation}
\mathcal{L} = \mathcal{L}_{task} + \lambda_{FR}\mathcal{L}_{FR} + \lambda_{AP}\mathcal{L}_{AP},
\end{equation}
where $\lambda_{FR}$ and $\lambda_{AP}$ balance the contributions of the two distillation losses. Focusing on the distillation component, the distillation loss can be written as $\mathcal{L}_{d} = \lambda_{FR}\mathcal{L}_{FR} + \lambda_{AP}\mathcal{L}_{AP}$.
In terms of mutual information, following Eq.~\ref{eq:loss_fr_IB} and Eq.~\ref{eq:loss_ap_IB}, this corresponds to
\begin{equation}
\max I_\alpha(F_a^T; F_a^S) - \frac{\lambda_{FR}}{\lambda_{AP}} I_\alpha(h_0^T, h_L^T; F_a^T),
\end{equation}
which enforces a compact representation that selectively preserves the most relevant alignment flow, and facilitates robust alignment transfer from teacher to student. This is consistent with the information bottleneck principle~\cite{oh2025visual}, retaining target-relevant information while discarding irrelevant input.

\section{Experiments}

\subsection{Experimental Setup}

\paragraph{Datasets}
To evaluate the effectiveness of LLaVAFlow, we follow prior work~\cite{liu2023visual,liu2024improved,liang2025lorasculpt} and fine-tune the model on Visual Question Answering (VQA) and Captioning tasks using the ScienceQA~\cite{lu2022learn} and COCO-Caption~\cite{lin2014microsoft} datasets. To assess general knowledge retention in the MLLM, we evaluate its performance on four upstream datasets after fine-tuning on the downstream tasks, including OKVQA~\cite{marino2019ok}, OCRVQA~\cite{mishra2019ocr}, GQA~\cite{hudson2019gqa}, and TextVQA~\cite{singh2019towards}. We report Accuracy for VQA tasks and CIDEr for Captioning.

\textbf{}{Implementation Details}
All compared baselines are adapted using LoRA and implemented within the LLaVA-1.5 framework~\cite{liu2024improved}. Following prior work~\cite{liu2024improved,liang2025lorasculpt}, we apply LoRA with ranks of 32 and 64 to all layers of the LLM using a learning rate of 2e-4, and perform full fine-tuning in the connector with a learning rate of 2e-5. Training is conducted for one epoch with a batch size of 16. All experiments are carried out on a single NVIDIA A100 80GB GPU. The key hyperparameters are set to $(\lambda_{FR}, \lambda_{AP}) = (1.5, 0.5)$ for VQA and $(1.5, 2.0)$ for Captioning.

\paragraph{Baselines}
We evaluate LLaVAFlow in a plug-and-play fashion, integrating it with existing state-of-the-art methods. This demonstrates its ability to preserve alignment flow, which consistently enhances both downstream performance and model generalization.
The evaluation covers a diverse set of baselines, including parameter-efficient fine-tuning methods (LoRA~\cite{hu2022lora}, DoRA~\cite{liu2024dora}), MLLM distillation approaches (LLaVAMod~\cite{shu2025llava}, LLaVAKD~\cite{cai2025llava}), post-training methods (DARE~\cite{yu2024language}, Model Tailor~\cite{zhu2024model}), and regularization strategies (LoRASculpt~\cite{liang2025lorasculpt}). Details of the compared baselines are provided in Appendix A.

\begin{figure*}[h]
  \centering
  \begin{subfigure}[b]{0.24\linewidth}
    \includegraphics[width=\linewidth]{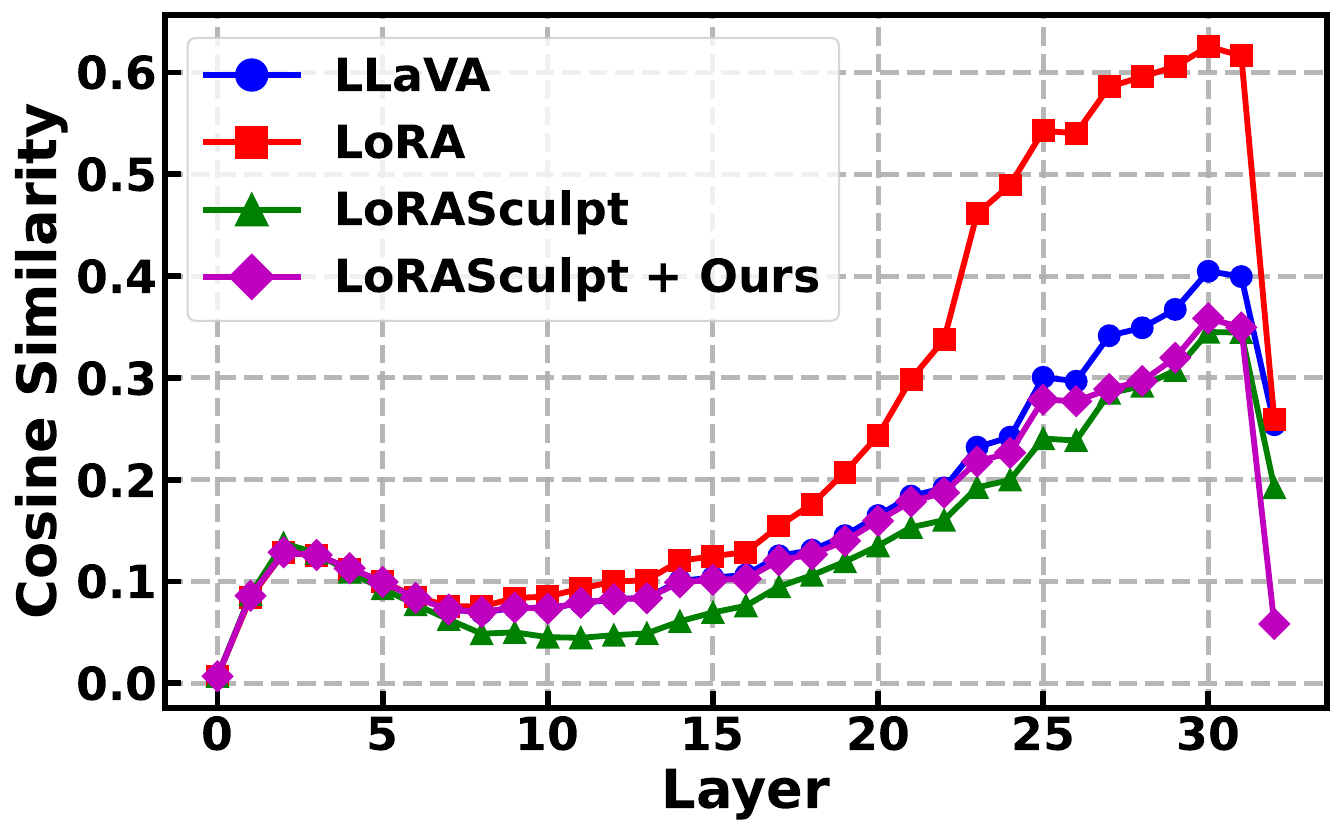}
    \caption{Cosine similarity on target}
    \label{fig:cosine_similarity_target}
  \end{subfigure}
  \begin{subfigure}[b]{0.24\linewidth}
    \includegraphics[width=\linewidth]{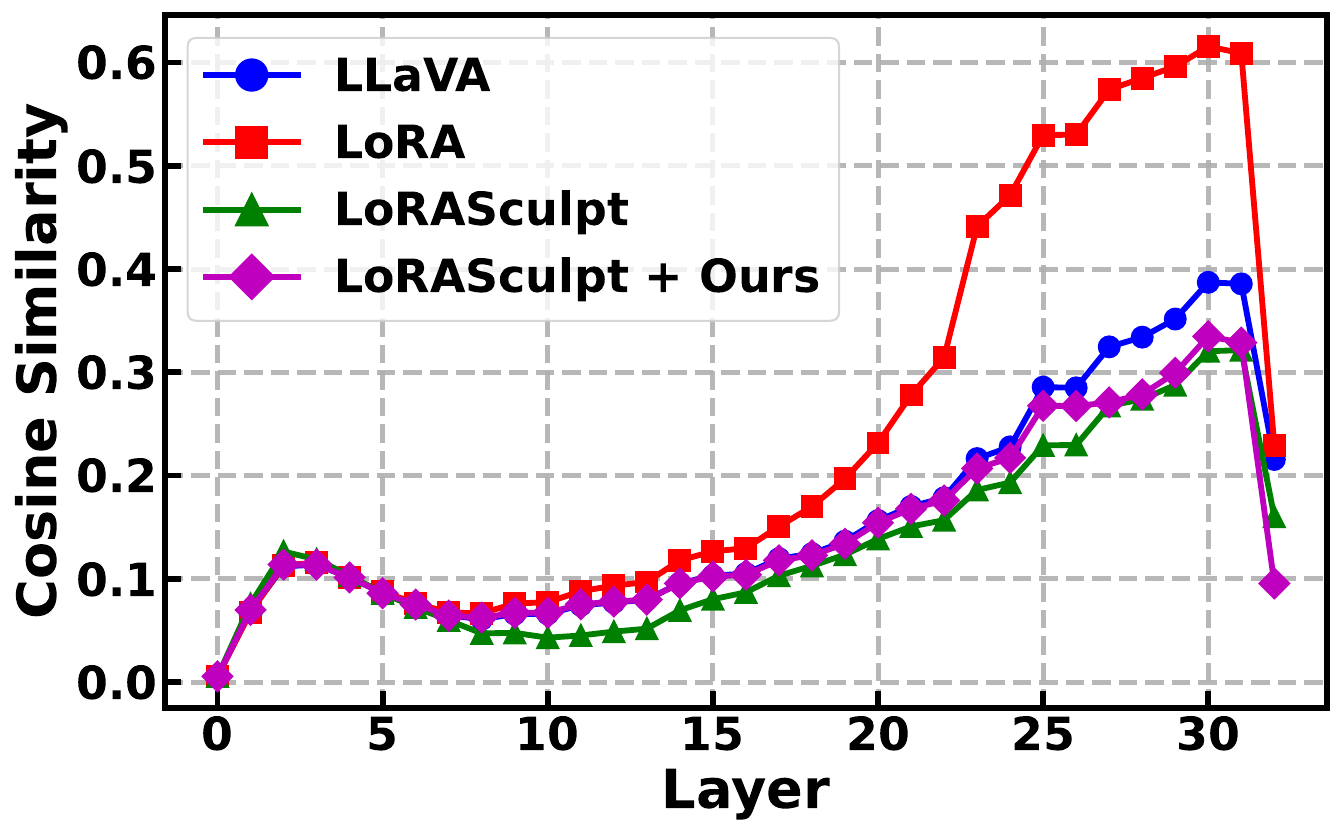}
    \caption{Cosine similarity on source}
    \label{fig:cosine_similarity_source}
  \end{subfigure}
  \begin{subfigure}[b]{0.24\linewidth}
    \includegraphics[width=\linewidth]{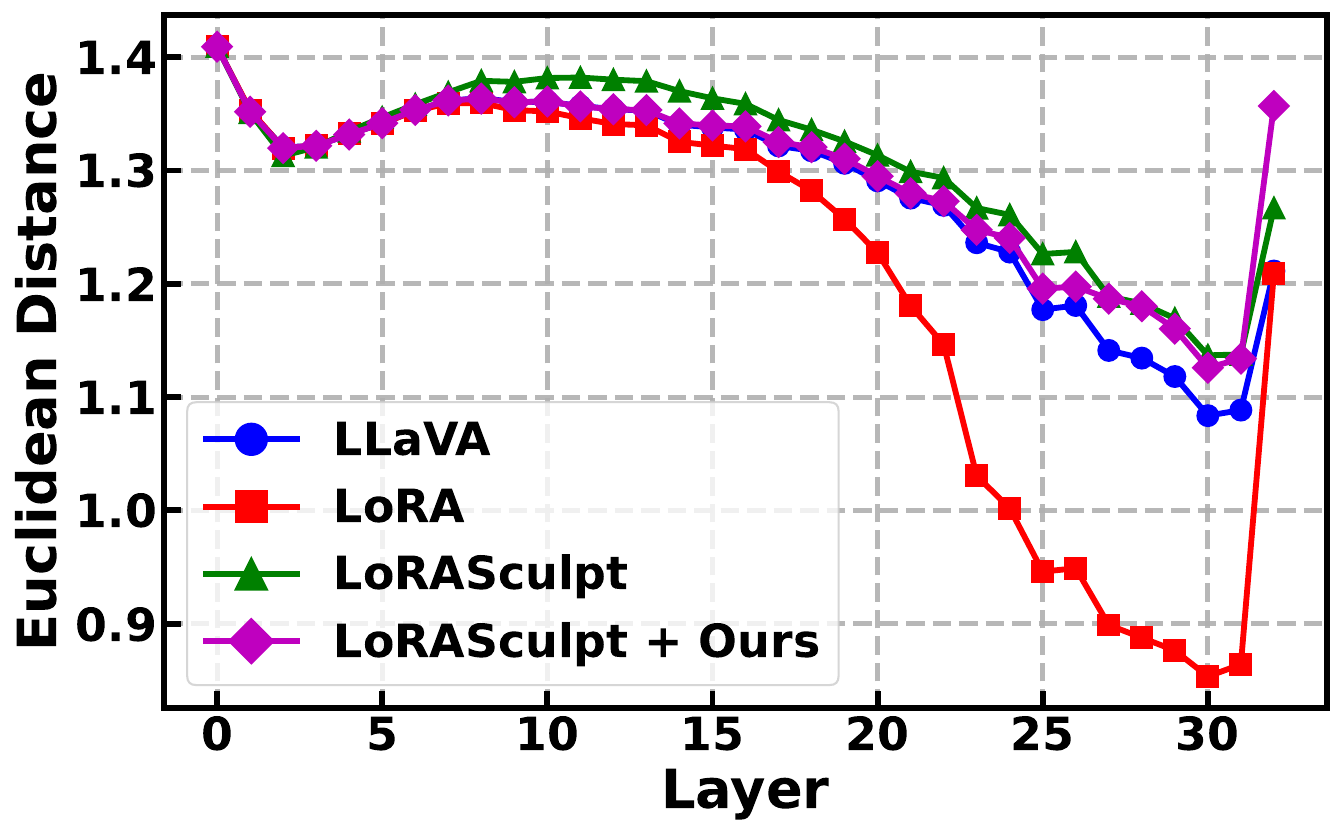}
    \caption{Euclidean distance on target}
    \label{fig:euclidean_distance_target}
  \end{subfigure}
  \begin{subfigure}[b]{0.24\linewidth}
    \includegraphics[width=\linewidth]{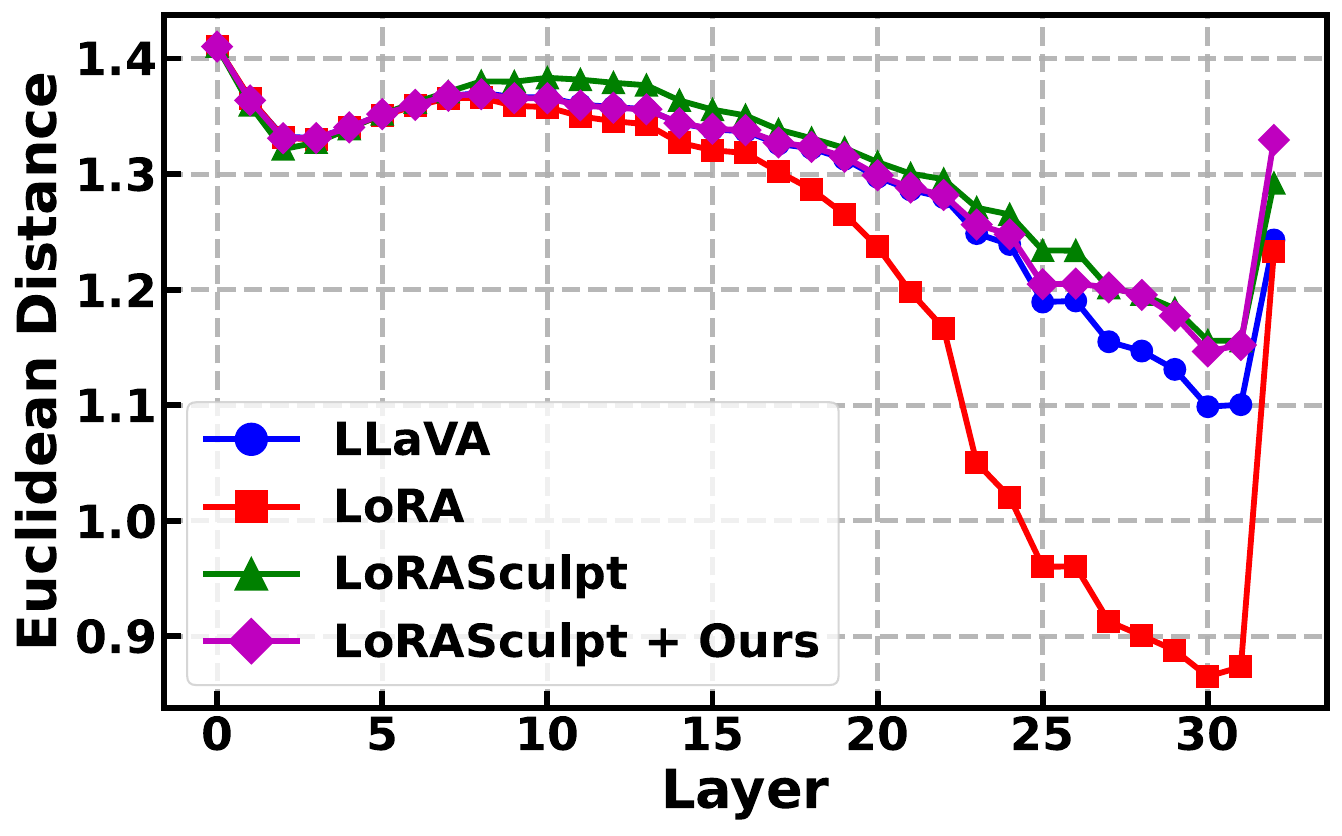}
    \caption{Euclidean distance on source}
    \label{fig:euclidean_distance_source}
  \end{subfigure}
  \caption{Alignment trends across layers. (a) and (b) show the average cosine similarity between features at the original vision and text embedding positions within each layer on the target and source datasets, respectively. (c) and (d) show the corresponding normalized Euclidean distance. All fine-tuned models are fine-tuned on ScienceQA~\cite{lu2022learn}. Our model preserves the pre-trained modality alignment in the shallow layers (1--25), indicating retention of general cross-modal alignment, while exhibiting divergence in the deeper layers as they adapt to task-specific objectives.}
  \label{fig:preservation_of_alignment}
\end{figure*}

\begin{figure*}[h]
    \centering
    \includegraphics[width=0.95\linewidth]{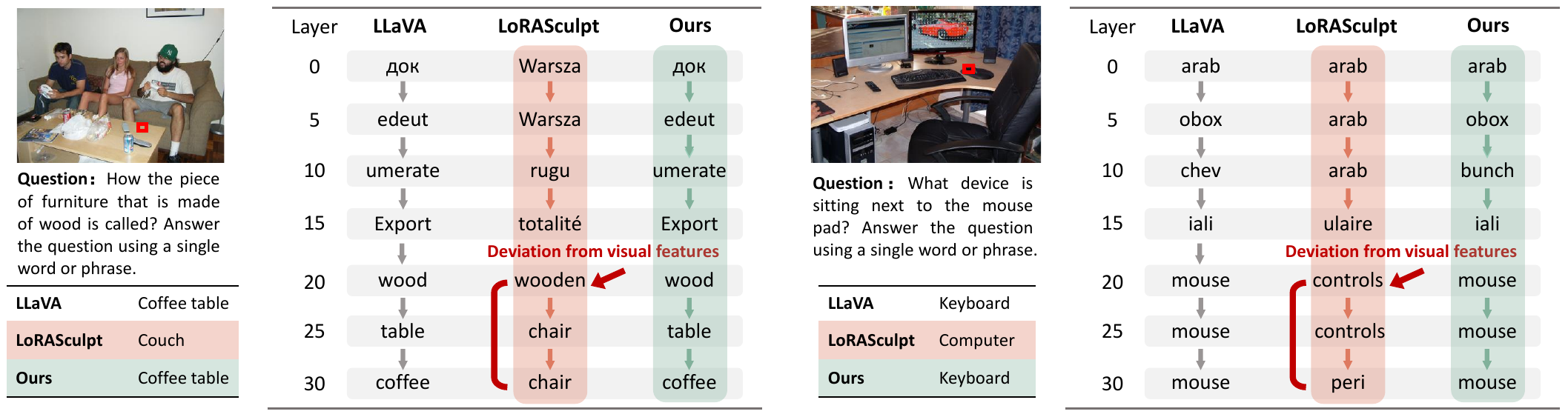}
    \caption{Comparison of semantic trajectories across models, obtained via Logit Lens~\cite{nostalgebraist2020logitlens}. Existing methods, such as LoRASculpt, fail to preserve alignment within the feature flow, leading to incorrect predictions. In contrast, our method retains most of the trajectory of the pre-trained LLaVA, effectively mitigating catastrophic forgetting. The trajectory is derived from the image patch indicated by the red box. Full trajectories are provided in Appendix B.1.}
    \label{fig:exp_trajectory}
\end{figure*}

\subsection{Main Results}
 
\paragraph{Comparison to State-of-the-Arts} 
We conduct extensive experiments by equipping LLaVAFlow with various state-of-the-art baselines across different rank settings on two downstream tasks. As shown in Table~\ref{tab:main_results}, LLaVAFlow brings notable gains on both Source and Target datasets, indicating that preserving alignment flow helps retain pretrained knowledge and facilitate downstream adaptation simultaneously. Importantly, these improvements hold across diverse methods, spanning adapter tuning (LoRA, DoRA), knowledge distillation (LLaVA-KD, LLaVA-Mod), and parameter regularization (DARE, Model Tailor, LoRASculpt), confirming the broad compatibility of our framework. This also suggests that alignment flow preservation offers a complementary benefit that is largely orthogonal to existing strategies, providing a new and effective angle for mitigating forgetting in MLLMs.

\subsection{Interpretation of LLaVAFlow}

\paragraph{Preservation of Alignment}

To investigate modality alignment, we analyze the alignment trends across layers by measuring the average cosine similarity and normalized Euclidean distance between features at the original vision and text embedding positions across different models. The experiments are conducted on ScienceQA~\cite{lu2022learn} (target dataset) and OCRVQA~\cite{mishra2019ocr} (source dataset), and all fine-tuned models are fine-tuned on ScienceQA~\cite{lu2022learn}.

As shown in Figure~\ref{fig:preservation_of_alignment}, in the shallow layers (1--25), the cosine similarity and normalized Euclidean distance of our model closely resemble those of the pre-trained LLaVA model. Since these layers primarily capture general knowledge~\cite{yosinski2014transferable,zhao2024layer}, maintaining the original cross-modal alignment is crucial for preserving the representational capacity acquired during pre-training~\cite{liang2022mind, feng2025align}. The consistent modality gap across these layers confirms that our method successfully retains this alignment. In the deeper layers (beyond layer 25), however, the modality gap of our model begins to diverge. This is reasonable, as the deeper layers are more responsible for encoding task-specific knowledge~\cite{yosinski2014transferable,zhao2024layer}, where the alignment naturally shifts to accommodate downstream objectives. This behavior reflects the strength of our information distillation strategy, which focuses on relational knowledge transfer rather than hard mimicry, enabling the model to preserve pre-trained alignment where it matters most while allowing sufficient flexibility for task adaptation.
This phenomenon also suggests that, although the alignment flow is computed using only the embeddings at the input and output of the network, it effectively regulates the entire forward feature propagation process.

\paragraph{Semantics Flow of Visual Tokens}
To demonstrate alignment flow preservation, we visualize the semantic trajectories of different methods by applying the LM head to intermediate hidden states, following the Logit Lens paradigm.

As shown in Figure~\ref{fig:exp_trajectory}, in Example 1, LoRASculpt loses critical vision-language alignment information along the trajectory and fails to map the image patch of \textit{cat} to the correct textual token \textit{cat}, unlike the pre-trained LLaVA model. In contrast, our method preserves most of the trajectory of LLaVA and produces the correct answer, indicating effective alignment flow preservation and mitigation of catastrophic forgetting.

\subsection{Ablation Studies}

\begin{table}[t]
\centering
\footnotesize
\setlength{\tabcolsep}{6pt}
\renewcommand{\arraystretch}{0.9}
\caption{Ablation study of Flow Refinement Loss~$\mathcal{L}_{FR}$ and Alignment Preservation Loss~$\mathcal{L}_{AP}$. All experiments are conducted with LoRA on the ScienceQA dataset.}
\label{tab:ablation_main}
\begin{tabular}{cc|ccc|ccc}
\toprule
\multirow{2}{*}{$\mathcal{L}_{FR}$} & \multirow{2}{*}{$\mathcal{L}_{AP}$} & \multicolumn{3}{c|}{\textit{LoRA Rank}~$=32$} & \multicolumn{3}{c}{\textit{LoRA Rank}~$=64$} \\
\cmidrule(lr){3-5} \cmidrule(lr){6-8}
 & & Target & Source & Avg & Target & Source & Avg \\
\midrule
\multicolumn{2}{c|}{Zero-shot} & 70.22 & 61.09 & 65.65 & 70.22 & 61.09 & 65.65 \\
\midrule
\ding{55} & \ding{55} & 87.57 & 60.10 & 73.84 & 88.49 & 55.67 & 72.08 \\
\ding{55} & \ding{51} & 89.08 & 60.04 & 74.56 & 87.86 & 58.71 & 73.28 \\
\ding{51} & \ding{55} & 89.37 & 60.05 & 74.71 & 88.52 & 57.11 & 72.81 \\
\ding{51} & \ding{51} & \textbf{89.46} & \textbf{60.21} & \textbf{74.84} & \textbf{89.46} & \textbf{58.13} & \textbf{73.79} \\
\bottomrule
\end{tabular}
\end{table}

\begin{table}[t]
\centering
\footnotesize
\setlength{\tabcolsep}{6pt}
\renewcommand{\arraystretch}{0.9}
\caption{Comparison of different alignment flow modeling strategies. All experiments are conducted with LoRA on the ScienceQA dataset. Attn-$L$ denotes stacking $L{-}1$ standard attention layers followed by AFM as the final scoring layer.}
\label{tab:ablation_alignment_model}
\begin{tabular}{l|ccc|ccc}
\toprule
\multirow{2}{*}{Method} & \multicolumn{3}{c|}{\textit{LoRA Rank}~$=32$} & \multicolumn{3}{c}{\textit{LoRA Rank}~$=64$} \\
\cmidrule(lr){2-4} \cmidrule(lr){5-7}
 & Target & Source & Avg & Target & Source & Avg \\
\midrule
Zero-shot   & 70.22 & 61.09 & 65.65 & 70.22 & 61.09 & 65.65 \\
\midrule
Scale     & 88.92 & 59.97 & 74.45 & 87.97 & 57.18 & 72.58 \\
Linear    & 88.73 & 59.34 & 74.03 & 87.55 & 34.99 & 61.27 \\
Attn-2    & 89.39 & 59.70 & 74.54 & 87.34 & 52.02 & 69.68 \\
Attn-3    & 88.23 & 60.12 & 74.18 & 86.68 & 38.61 & 62.64 \\
Ours      & \textbf{89.46} & \textbf{60.21} & \textbf{74.84} & \textbf{89.46} & \textbf{58.13} & \textbf{73.79} \\
\bottomrule
\end{tabular}
\end{table}

\paragraph{Ablation of Information Bottleneck Distillation}
We conduct an ablation study to evaluate the contribution of each component in our proposed LLaVAFlow: Flow Refinement Loss~$\mathcal{L}_{FR}$ and Alignment Preservation Loss~$\mathcal{L}_{AP}$. Experiments are performed on the ScienceQA dataset using LoRA with ranks of 32 and 64. To ablate $\mathcal{L}_{FR}$, we replace the learnable relation module with a fixed dot-product operation and remove the Flow Refinement Loss. To ablate $\mathcal{L}_{AP}$, we substitute the information-based loss with an MSE loss.
As shown in Table~\ref{tab:ablation_main}, both components contribute positively to the overall performance. 
Notably, under rank 64, the LoRA baseline suffers from a significant drop in source performance (55.67), indicating catastrophic forgetting. In this setting, $\mathcal{L}_{AP}$ alone recovers the source performance from 55.67 to 58.71, demonstrating its effectiveness in preserving the pre-trained alignment flow. Meanwhile, $\mathcal{L}_{FR}$ contributes more to target task adaptation, boosting target accuracy from 88.49 to 88.52, suggesting that the information-bottleneck-based loss effectively filters out noisy signals that would otherwise harm downstream performance. When both losses are combined, the model achieves the highest target accuracy (89.46) while substantially mitigating catastrophic forgetting on the source datasets (58.13), confirming that the two components are complementary.

\paragraph{Ablation of Alignment Flow Module Design} 
We conduct a series of experiments on the ScienceQA dataset using LoRA with a rank of 32 to evaluate the design choices of our Alignment Flow Module (AFM). We compare the proposed AFM against three alternative architectures for modeling alignment flow between $h_0$ and $h_L$, i.e., the embeddings before the first layer and after the final layer of the LLM backbone: (1)~\textbf{Scale Module}, which applies learnable channel-wise scaling factors to compute relation scores between $h_0$ and $h_L$ via a scaled dot product, augmented with a learnable bias term; (2)~\textbf{Linear Module}, which constructs pairwise feature concatenations across all position pairs of $h_0$ and $h_L$ and maps them to relation scores through a linear layer, combined with residual dot-product scores; (3)~\textbf{Attention Module ($L$ layers)}, which stacks $L{-}1$ standard attention layers with residual connections to iteratively refine $h_L$ by attending to $h_0$, followed by our AFM as the final scoring layer. We evaluate $L=2$ and $L=3$.
As shown in Table~\ref{tab:ablation_alignment_model}, our single-layer AFM achieves the best overall performance under both rank settings. The differences among methods are relatively small at rank 32 but become substantially larger at rank 64, where the multi-layer Attention modules suffer severe source degradation, with source accuracy dropping to as low as 38.61. We attribute this to the over-parameterized projection modules absorbing the teacher's alignment patterns within their own parameters~\cite{chen2022improved}, rather than forcing the student backbone to internalize this knowledge. This weakens the implicit regularization provided by the distillation objective, allowing the LoRA parameters to be dominated by the downstream task loss and resulting in catastrophic forgetting. The problem is amplified at rank 64 due to the increased degrees of freedom. In contrast, our lightweight single-layer AFM serves as an effective conduit for learning alignment knowledge and transferring it to the student, thereby achieving a better balance between task adaptation and alignment retention.

\begin{figure}[t]
\centering
\captionsetup[subfigure]{font=footnotesize}

\begin{subfigure}{0.24\linewidth}
    \centering
    \includegraphics[width=\linewidth]{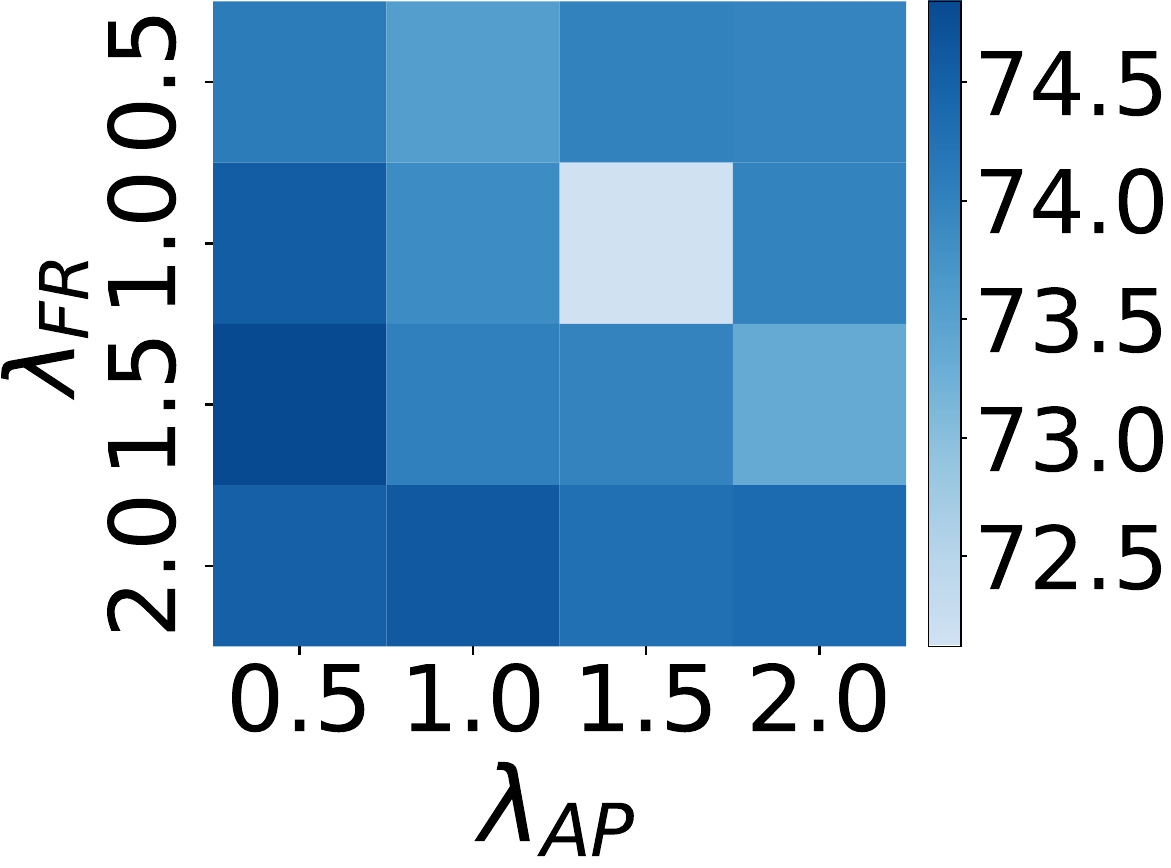}
    \caption{LoRA, VQA}
\end{subfigure}
\hfill
\begin{subfigure}{0.24\linewidth}
    \centering
    \includegraphics[width=\linewidth]{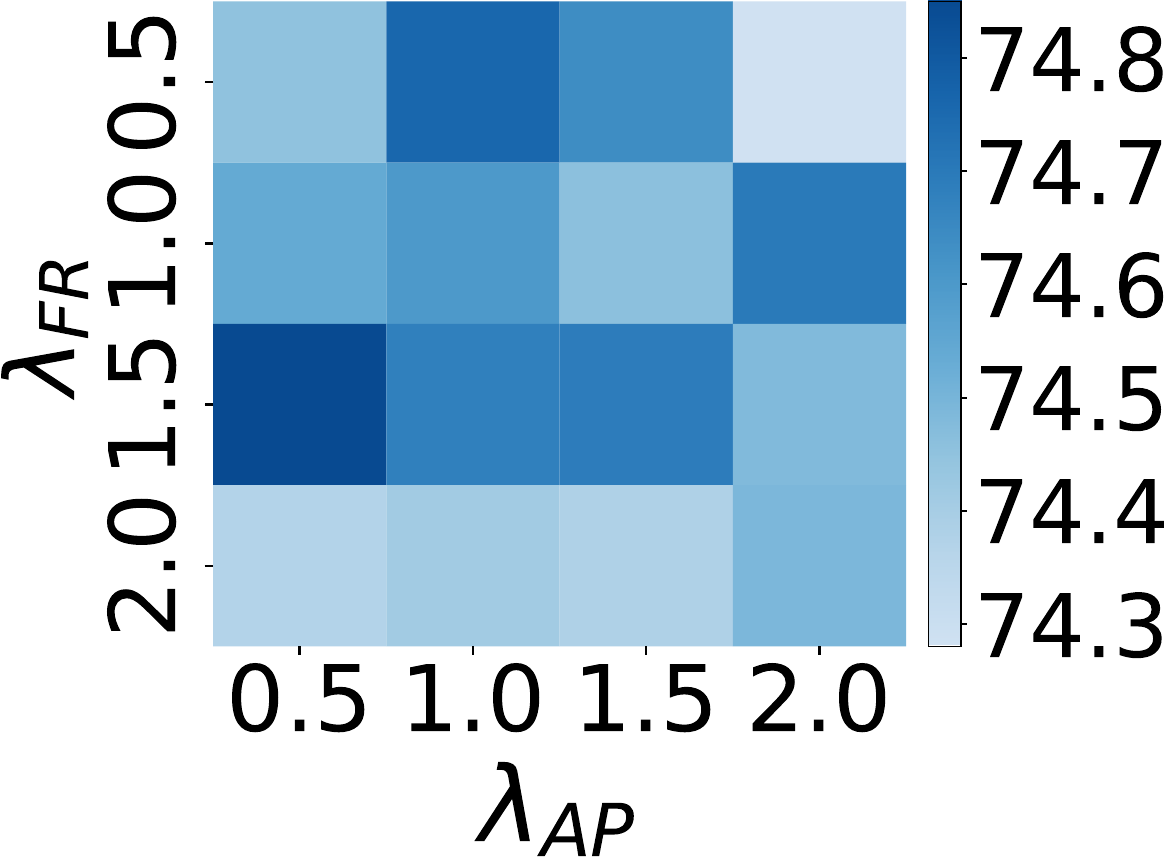}
    \caption{LLaVAKD, VQA}
\end{subfigure}
\hfill
\begin{subfigure}{0.24\linewidth}
    \centering
    \includegraphics[width=\linewidth]{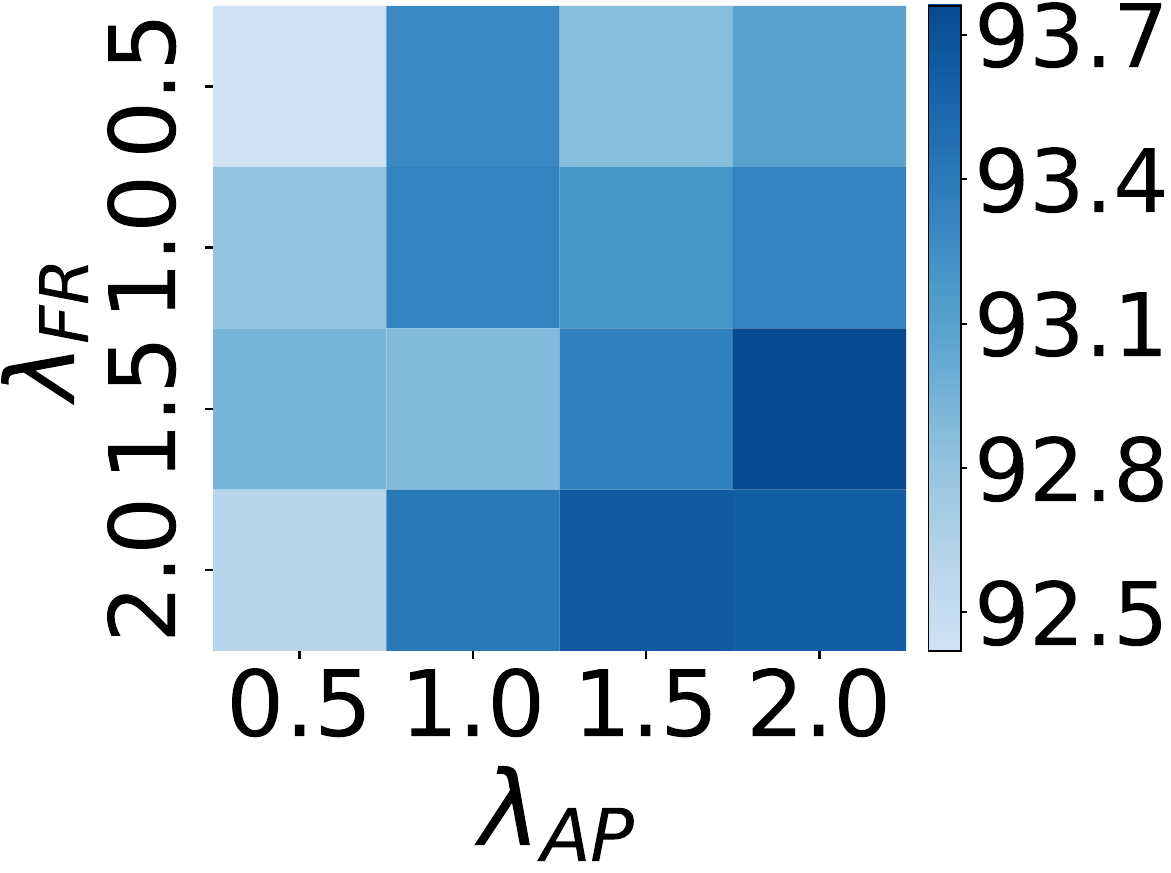}
    \caption{LoRA, Cap.}
\end{subfigure}
\hfill
\begin{subfigure}{0.24\linewidth}
    \centering
    \includegraphics[width=\linewidth]{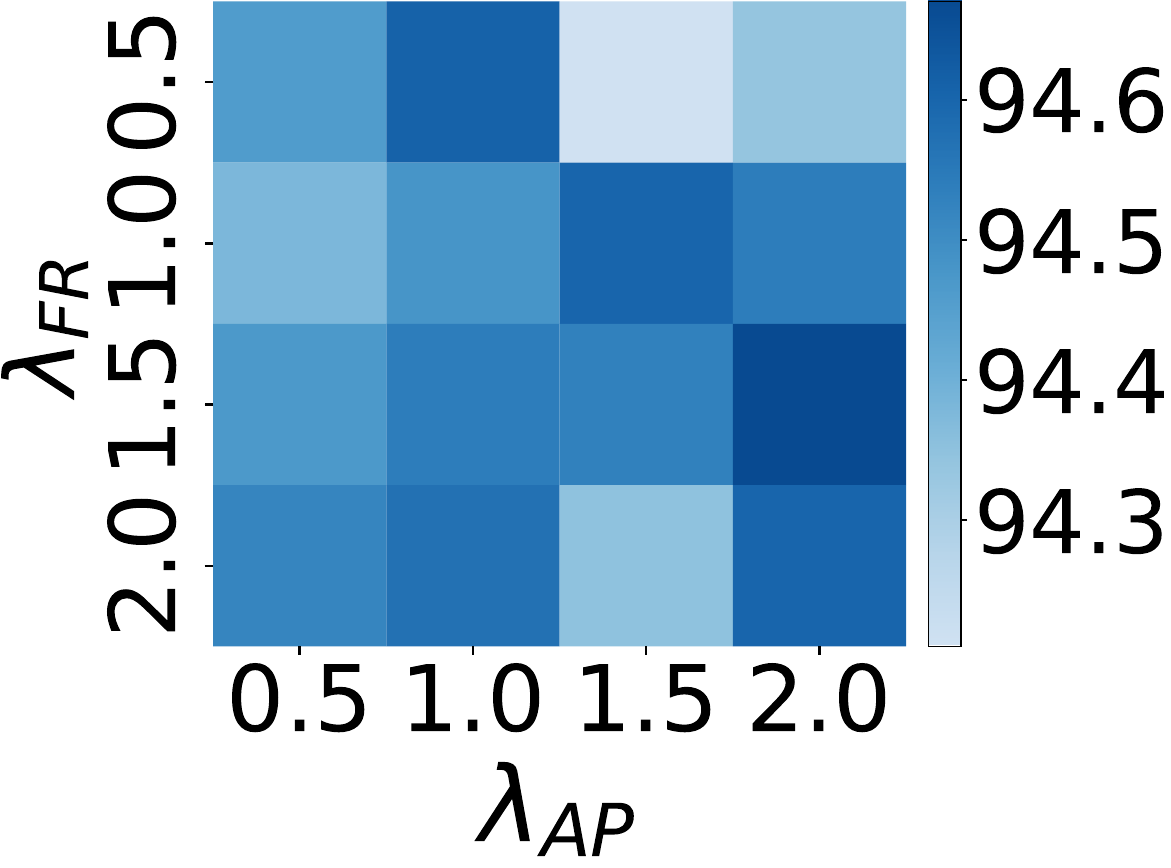}
    \caption{LLaVAKD, Cap.}
\end{subfigure}

\caption{Hyperparameter analysis of $\lambda_{FR}$ and $\lambda_{AP}$. All results are obtained with LLaVAFlow applied.}
\label{fig:ablation_hyperparameter}
\end{figure}

\paragraph{Analysis of Hyperparameters}

We conduct a sensitivity analysis of the hyperparameters $\lambda_{FR}$ and $\lambda_{AP}$, which regulate the trade-off between the Flow Refinement loss $\mathcal{L}_{FR}$ and the Alignment Preservation loss $\mathcal{L}_{AP}$ across diverse settings. Specifically, we evaluate their effects on two tasks, VQA and Captioning, and integrate our method with two representative paradigms: LoRA as a parameter-efficient fine-tuning approach, and LLaVA-KD as a distillation-based method.
As shown in Figure~\ref{fig:ablation_hyperparameter}, each subfigure presents a heatmap of the average performance over both source and target datasets under different hyperparameter configurations.
Based on these empirical observations, we select $\lambda_{FR}=1.5, \lambda_{AP}=0.5$ for VQA, and $\lambda_{FR}=1.5, \lambda_{AP}=2.0$ for Captioning.

\section{Conclusion}

In this paper, we present LLaVAFlow, an information-theoretic distillation framework designed to mitigate catastrophic forgetting in MLLMs during visual instruction tuning. We introduce the concept of alignment flow, which captures cross-modal alignment information implicitly encoded in the feature propagation trajectory of the LLM backbone.
To extract this alignment flow, we design the Alignment Flow Module (AFM) as a lightweight relational module, and employ an information bottleneck objective that compresses the mutual information between the AFM output and the raw MLLM embeddings, thereby encouraging the extraction of alignment-relevant signals while discarding redundant noise. To transfer the extracted alignment flow, we maximize the mutual information between the alignment flows of the pretrained and fine-tuned models, preserving structural cross-modal knowledge throughout adaptation.
Extensive experiments on VQA and Captioning tasks demonstrate that LLaVAFlow functions as a plug-and-play framework, achieving strong downstream performance while effectively retaining the general knowledge of the MLLM.

\begin{acks}
This work was supported in part by the New Generation Artificial Intelligence-National Science and Technology Major Project No. 2025ZD0123003, NSFC under Grant 62576268 and 62302384, and the Key Research and Development Project in Shaanxi Province No. 2024PT-ZCK-89.
\end{acks}

\bibliographystyle{ACM-Reference-Format}
\balance
\bibliography{refs}


\end{document}